\documentclass[]{servicenow} 

\usepackage{amsmath,amsfonts,bm}

\def\eqref#1{equation~\ref{#1}}

\def\1{\bm{1}}

\DeclareMathAlphabet{\mathsfit}{\encodingdefault}{\sfdefault}{m}{sl}
\SetMathAlphabet{\mathsfit}{bold}{\encodingdefault}{\sfdefault}{bx}{n}

\usepackage{hyperref}
\usepackage[inline]{enumitem}
\usepackage{graphicx}
\usepackage{url}
\usepackage{subfig}
\usepackage{nicefrac}
\usepackage{cleveref}
\usepackage[table]{xcolor} 
\usepackage{array}         
\usepackage{booktabs}      

\usepackage{microtype}
\usepackage{enumitem}
\usepackage{colortbl}
\usepackage{xcolor}
\usepackage{multirow}
\usepackage{wrapfig}
\usepackage{tikz}
\usepackage{pgfplots}
\usepackage{tcolorbox}
\usepackage{tabularx}
\usepackage{alltt}
\tcbuselibrary{breakable, skins}
\pgfplotsset{compat=1.18}
\usetikzlibrary{patterns, fillbetween}
\usepgfplotslibrary{groupplots}
\usetikzlibrary{arrows.meta,positioning,fit,backgrounds,calc,decorations.pathreplacing}
\newtcolorbox{taskbox}[1][]{
    colback=blue!2,
    colframe=blue!40,
    boxrule=0.5pt,
    arc=3pt,
    title=#1,
    coltitle=white,
    colbacktitle=blue!60,
    fonttitle=\bfseries\sffamily,
    breakable,
    enhanced,
    before skip=10pt,
    after skip=10pt,
    left=8pt, right=8pt,
    top=6pt, bottom=6pt,
}

\newtcolorbox{patternbox}[1][]{
    colback=blue!4,
    colframe=blue!30,
    boxrule=0.4pt,
    arc=2pt,
    left=10pt, right=10pt,
    top=4pt, bottom=4pt,
    before skip=2pt,
    after skip=2pt,
}

\definecolor{barblue}{HTML}{2171B5}
\definecolor{barorange}{HTML}{D94801}

\definecolor{frontierbg}{HTML}{E8F6FB}
\definecolor{basebg}{HTML}{FCECDF}
\definecolor{finetunedbg}{HTML}{EAF6E8}

\usepgfplotslibrary{groupplots}
\pgfplotsset{compat=1.18}
\usetikzlibrary{arrows.meta,positioning,fit,backgrounds,calc,decorations.pathreplacing}
\usepackage{hyperref}

\title{Do Enterprise Systems Need Learned World Models?
The Importance of Context to Infer Dynamics}

\author[1,*]{Jishnu Sethumadhavan Nair}
\author[1,*]{Patrice Bechard}
\author[1,*]{Rishabh Maheshwary}
\author[1,\dagger]{\\Surajit Dasgupta}
\author[1,\dagger]{Sravan Ramachandran}
\author[1,\dagger]{Aakash Bhagat}
\author[1,\dagger]{Shruthan Radhakrishna}
\author[1]{\\Pulkit Pattnaik}
\author[2]{Johan Obando-Ceron}
\author[1]{Shiva Krishna Reddy Malay}
\author[1]{\\Sagar Davasam}
\author[1]{Seganrasan Subramanian}
\author[1]{Vipul Mittal}
\author[1]{\\Sridhar Krishna Nemala}
\author[1,2]{Christopher Pal}
\author[1]{Srinivas Sunkara}
\author[1,2]{Sai Rajeswar}

\affiliation[1]{ServiceNow}
\affiliation[2]{Mila}

\contribution[*]{Co-first authors; contributed equally.}
\contribution[\dagger]{Co-second authors; contributed equally.}

\abstract{
    World models enable agents to anticipate the effects of their actions by internalizing environment dynamics. In enterprise systems, however, these dynamics are often defined by tenant-specific business logic that varies across deployments and evolves over time, making models trained on historical transitions brittle under deployment shift. We ask a question the world-models literature has not addressed: \emph{when the rules can be read at inference time, does an agent still need to learn them?} We argue, and demonstrate empirically, that in settings where transition dynamics are configurable and readable, runtime discovery complements offline training by grounding predictions in the active system instance. We propose \emph{enterprise discovery agents}, which recover relevant transition dynamics at runtime by reading the system's configuration rather than relying solely on internalized representations. We introduce \emph{CascadeBench}, a reasoning-focused benchmark for enterprise cascade prediction that adopts the evaluation methodology of World of Workflows on diverse synthetic environments, and use it together with deployment-shift evaluation to show that offline-trained world models can perform well in-distribution but degrade as dynamics change, whereas discovery-based agents are more robust under shift by grounding their predictions in the current instance. Our findings suggest that, in configurable enterprise environments, agents should not rely solely on fixed internalized dynamics, but should incorporate mechanisms for discovering relevant transition logic at runtime.
}

\correspondence{\email{jishnus.nair@servicenow.com}, \email{sai.mudumba@servicenow.com}}

\begin{document}

\maketitle

\section{Introduction}

Large Language Model (LLM) agents~\citep{yao2022react, wang2024survey} are increasingly deployed in environments with complex dynamics. To plan and act effectively over long horizons, these agents must understand how their actions affect the environment, enabling accurate anticipation of downstream state changes~\citep{erdogan2025planandact,gu2025webdreamer}. This ability to capture environment dynamics, whether implicitly or explicitly, is central to building reliable autonomous agents in enterprise settings~\citep{gupta2026world}.

Enterprise systems differ from traditional environments because their dynamics are partly specified by tenant-specific configuration artifacts, such as business rules and workflows, that vary across deployments and evolve over time~\citep{bezemer2010multitenant,makki2018multitenant}. Thus, the same action can have different effects depending on the active configuration of the current system instance. Learned \emph{enterprise world models} can capture recurring patterns within a fixed deployment or workflow family, but models trained only on historical transitions may become brittle under deployment shift~\citep{doshivelez2016hidden,lee2020contextaware}.

This raises an alternative: instead of internalizing dynamics ahead of time, agents can \emph{discover} them at runtime. We define \emph{enterprise discovery agents} as agents that actively recover transition logic by interacting with the system (e.g. by querying state, inspecting workflow definitions, or issuing targeted probe actions). This strategy is natural in enterprise systems, where transition logic is often exposed through configuration artifacts such as business rules and workflows. Our comparison asks whether agents should rely solely on internalized dynamics when the rules governing the current environment can be inspected directly.

\begin{figure}[t]
    \centering
    \includegraphics[width=\linewidth]{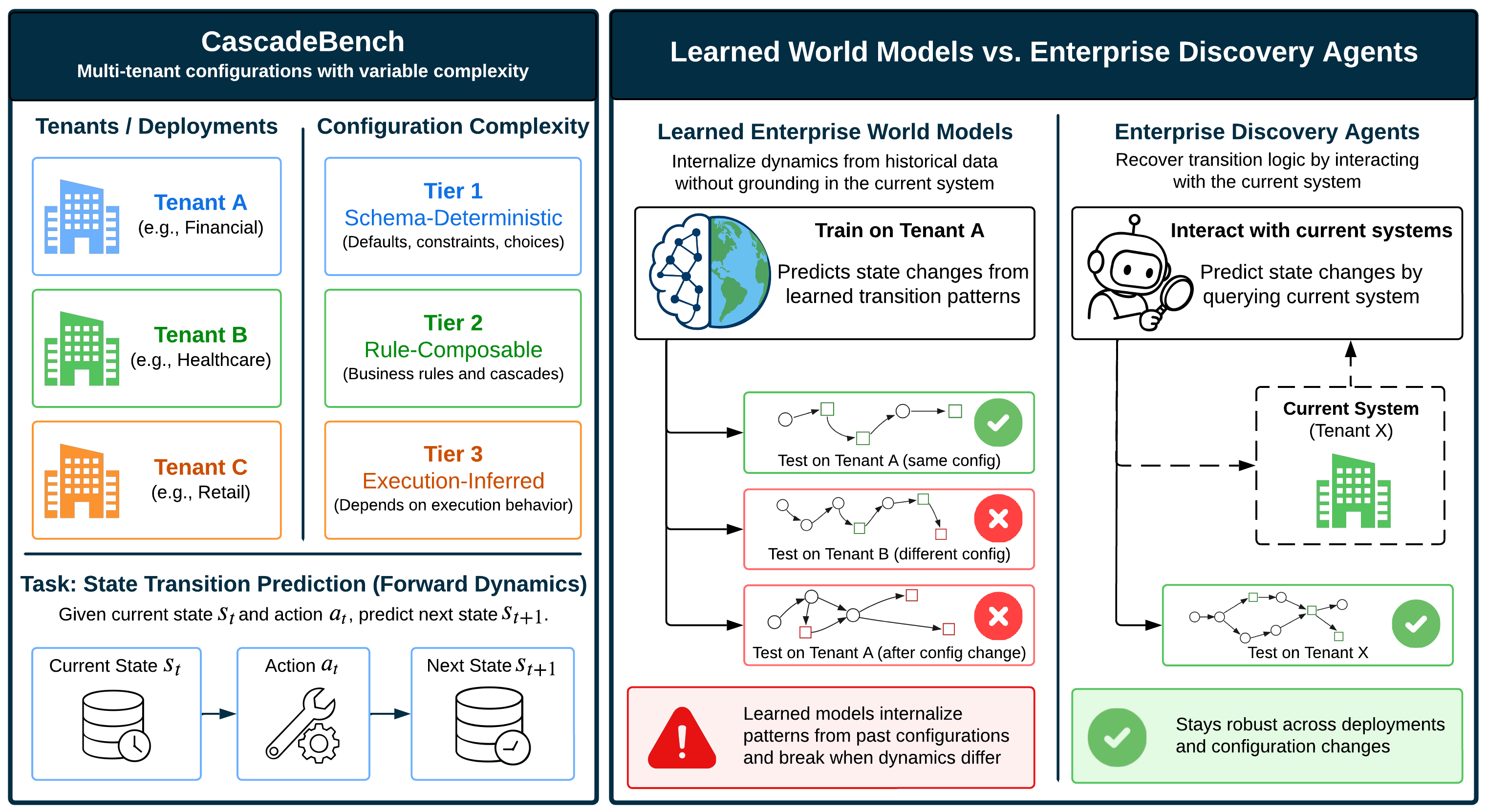}
    \vspace{-17pt}
    \caption{
    \textbf{Overview of CascadeBench and comparison between learned world models and enterprise discovery agents.}
    \emph{Left}: CascadeBench evaluates agents across multi-tenant enterprise environments and varying configuration complexity tiers. Task requires predicting the next state $s_{t+1}$ given the current state $s_t$ and action $a_t$.
    \emph{Right}: Learned enterprise world models internalize transition dynamics from historical data, performing well in-distribution but degrading under deployment and configuration shift when not grounded in the active instance. Enterprise discovery agents interact with the current system (e.g., querying state and inspecting workflow definitions) to recover transition logic at runtime, enabling robust predictions across tenants and evolving configurations.
    }
    \vspace{-10pt}
    \label{fig:teaser}
\end{figure}

We evaluate this question on \emph{CascadeBench}, a benchmark for enterprise cascade prediction under configuration and deployment shift. We show that offline-trained world models perform well in-distribution but degrade as dynamics change, while discovery agents remain more robust by grounding predictions in the active deployment. These results suggest that enterprise world modeling should combine learned priors with runtime discovery rather than rely only on fixed internalized dynamics.

\begin{tcolorbox}[colback=orange!15,
leftrule=0.5mm,top=0.5mm,bottom=0.5mm]
\textbf{Our contributions are as follows:}

\begin{enumerate}[left=0.3cm,nosep]
    \item We formalize enterprise dynamics as deployment-specific and evolving transition functions defined by system configuration.
    \item We introduce \emph{CascadeBench}, a benchmark for evaluating robustness under configuration and deployment shift across varying levels of transition complexity.
    \item We propose \emph{enterprise discovery agents}, which recover dynamics at runtime through interaction with the system, improving robustness under shift.
    \item We show that offline-trained enterprise world models perform well in-distribution but degrade under configuration shift when not grounded in the active deployment.
\end{enumerate}
\end{tcolorbox}

\section{Related Work}
\label{sec:related}

\paragraph{World models for decision-making agents.}
World models aim to enable agents to anticipate the effects of their actions by learning environment dynamics~\citep{hafner2019learning,hafner2020dream,hansen2024tdmpc}. Early work by \citet{schmidhuber1990making} introduced the idea of separating a predictive model and control, forming the foundation of model-based reinforcement learning. This paradigm has since been extended by methods such as World Models~\citep{ha2018world} and Dreamer~\citep{hafner2020dream,hafner2025mastering}, which learn latent dynamics to support planning and policy optimization. Later work improves scalability through better latent representations, longer-horizon rollouts, and tighter integration between planning and learning~\citep{hansen2024tdmpc}. In visual and robotic settings, approaches such as the I-JEPA~\citep{assran2023self} and V-JEPA~\citep{assran2025v} similarly motivate learning predictive representations rather than predicting pixels directly.

More recently, world models have been adapted to language-based agents, where reasoning is framed as planning over simulated trajectories~\citep{hao-etal-2023-reasoning}. In these settings, the environment is a structured interface such as the web, code execution environments, or tool APIs. Methods such as WebDreamer~\citep{gu2025webdreamer}, Code World Models~\citep{copet2025cwm}, and Generative Tool Models (GTM)~\citep{ren2025gtm} learn to approximate environment responses, enabling agents to simulate interactions without executing them. Across these approaches, the common assumption is that environment dynamics should be internalized into a learned simulator. We study a complementary regime where system behavior is externally accessible at inference time through structured interfaces, logs, or configuration files. In such settings, learned simulators may introduce unnecessary approximation error and reduce robustness under distribution shift.

\paragraph{Agents interacting with structured environments.}
A complementary line of work studies agents that interact directly with environments to retrieve information or execute actions. Tool-augmented agents~\citep{yao2022react, schick2023toolformer} use external APIs and structured interfaces to ground reasoning in real system responses. Recent work shows that such agents can operate effectively in enterprise environments by querying platform APIs at runtime, avoiding the need to approximate system behavior~\citep{bechard2026terminal}. Beyond tool use, interaction can also serve as a mechanism for structure discovery. Agents can acquire reusable skills through exploration~\citep{wang2024voyager}, infer abstractions from structured interfaces~\citep{prabhu2026walt}, and recover latent environment dynamics through experimentation~\citep{jansen2024discoveryworld}. These approaches suggest that interaction provides a reliable and adaptive signal for understanding environment behavior, particularly in non-stationary or partially observable settings. Our work builds on this perspective by studying agents that explicitly recover transition dynamics from live system configurations, enabling robust behavior under distribution shift rather than relying solely on learned simulators.

\paragraph{Enterprise agent benchmarks.}
Existing enterprise benchmarks evaluate agents on task execution across UI- and API-based settings. UI-centric benchmarks such as WorkArena and WorkArena++~\citep{drouin2024workarena, boisvert2024workarena} focus on browser interaction with platforms like ServiceNow, exposing challenges in long-horizon planning, delayed feedback, and error accumulation. API-based benchmarks such as CRMArena~\citep{huang2024crmarena} operate over structured Salesforce environments, enabling more controlled evaluation but often restricting the action space and system complexity. Multi-domain settings like EnterpriseOps-Gym~\citep{malay2026enterpriseops} and TheAgentCompany~\citep{xu2026theagentcompany} expand coverage across enterprise tools and workflows, though they primarily emphasize task execution rather than understanding system dynamics.

World of Workflows (WoW)~\citep{gupta2026world} takes a different angle, evaluating agents' ability to predict state transitions, action effects, and constraints in enterprise workflows, showing that frontier models struggle with multi-step dynamics. However, WoW evaluates fixed configurations in zero-shot settings, leaving open how agents adapt when dynamics vary across deployments. We address this gap with CascadeBench, a reasoning-focused benchmark that adopts WoW's transition-prediction methodology on synthetic schemas designed to isolate reasoning from parametric memorization and retrieval noise. Rather than measuring prediction accuracy under fixed configurations, we study how agents recover and adapt to dynamics at inference time.

\section{Enterprise Dynamics}
\label{sec:dynamics}


An enterprise platform typically maintains a structured state encoded across interconnected database tables, which may include information such as: users, configuration items, incidents, changes, and Service Level Agreements (SLAs). An agent interacts with this state through API actions to create records, update fields, trigger workflows, etc. The consequences of any action depend not only on the current state and the action itself, but on a layer of customer-specific configuration that governs how the platform responds. We formalize this as a contextual transition model. Let $s_t$ denote the observable platform state at step $t$ (the set of record field values across all relevant tables), $a_t$ the action taken, and $c$ the instance configuration (the collection of all business rules\footnote{Abbreviated BR throughout. See Appendix~\ref{app:glossary} for ServiceNow-specific terminology used in this paper.}, workflow definitions, approval policies, SLA definitions, and access control lists deployed on a particular customer's instance),
\begin{equation}
    s_{t+1} \sim P(s_{t+1} \mid s_t, a_t, c).
\end{equation}
In standard world model settings, $c$ is fixed and unknown, and the agent must learn dynamics from interaction alone. Enterprise systems differ from standard world model settings in two ways. First, $c$ is not fixed. Administrators continuously modify rules, so dynamics shift without changes to the underlying platform. Second, $c$ is explicit and readable. Rules, workflows, and policies are stored as inspectable records with defined conditions and actions. The central question is whether a learned world model trained on transition data can reliably predict $s_{t+1}$ on its own, or whether accurate prediction requires runtime grounding in the active configuration $c$.
%
Furthermore, unlike formulations that model the full environment state, enterprise world models benefit from a sparse transition view. In practice, we effectively model a state delta, $\Delta s_t$, roughly corresponding to $s_{t+1} - s_t$: the subset of fields whose values change after action $a_t$. This focuses modeling capacity on the task-relevant parts of the enterprise state affected by the transition.

Actions compose through cascades: a single field update can induce chains of business rule executions that propagate across tables, initiate SLA timers, and schedule notifications. The resulting transition from $s_t$ to $s_{t+1}$ may therefore involve dozens of intermediate steps, with depth and branching determined entirely by the instance-specific configuration of interacting rules.

Not all state transitions are equally hard to predict. To clarify the sources of difficulty in this setting, we distinguish three levels of transition complexity: \emph{Tier~1} schema-determined effects, \emph{Tier~2} rule-composed cascades, and \emph{Tier~3} execution-inferred behavior. Table~\ref{tab:complexity_tiers} in the Appendix summarizes this taxonomy with concrete examples.
We use these tiers both to structure the benchmark and to scope our comparison. Tier~1 and Tier~2 transitions are recoverable, in principle, from inspectable configuration: schemas capture defaults and constraints, while active business rules capture multi-step cascades. Tier~3 transitions are only partially recoverable: the rules are still inspectable, but the realized outcome also depends on execution-order resolution and other engine-internal behaviors not exposed in static artifacts. We therefore treat Tier~3 as a partial structural limit rather than a hard ceiling, and report tier-stratified results separately.
\section{Enterprise Gym}
\label{sec:gym}

We define a world as $W = (E, T)$, where $E$ specifies the environment (organizational structure, configuration database, business rules, initial records, etc.) and $T$ is the transition function induced by $E$ on the platform. $T$ is not simulated: we deploy $E$ to a live platform instance so that when an agent acts, the real engine executes server-side scripts and the resulting state $s'$ is the actual database state, avoiding the simulation-to-production gap of approximated benchmarks.

\paragraph{Diversity at scale.}
Worlds are generated from a catalog of 1{,}596 business rule 
patterns spanning 6 industries and 11 operational domains, with each 
world instantiating a unique subset. A dependency-ordered construction 
pipeline expands ${\sim}$27{,}000 LLM-generated base scenarios into 
${\sim}$802{,}000 validated initial states. Diversity mechanisms, 
controlled rule-conflict injection, and validation guardrails are 
described in Appendix~\ref{app:gym}.

\paragraph{Data Collection.}
\begin{figure*}[t]
    \centering
    \includegraphics[width=\textwidth]{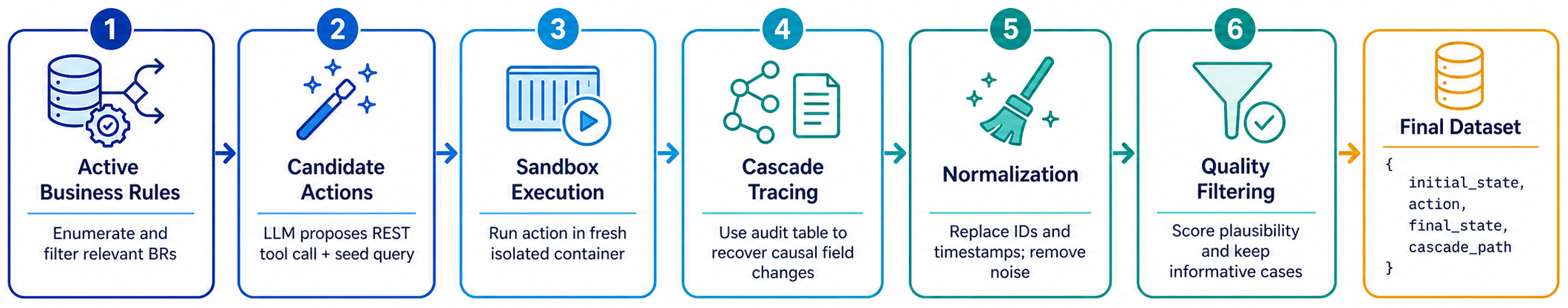}
    \vspace{-1.8em}
    \caption{\textbf{Pipeline for constructing the Business Rule cascade dataset.} Candidate actions are executed in isolated sandboxes, traced through audit logs, normalized, and filtered into a dataset of state transitions and cascade paths.}
    \label{fig:pipeline}
    \vspace{-12pt}
\end{figure*}

We collect ground-truth transition data by firing tool calls against the live worlds described above and recording the resulting cascades through platform audit logs. Figure~\ref{fig:pipeline} summarizes the pipeline: candidate tool calls are executed in isolated sandboxes, causal state changes are recovered from \texttt{sys\_audit}, platform-specific identifiers and noise are normalized away, and low-quality traces are filtered before inclusion. Each retained sample is a tuple $(s_t, a_t, s_{t+1}, \pi)$, where $s_t$ is the relevant initial state, $a_t$ is the executed tool call, $s_{t+1}$ is the post-execution state diff, and $\pi$ is the cascade path: the ordered sequence of tables touched and business rules attributed to each transition. The platform engine is the source of truth: when an action is fired, real business rules execute, real SLA timers start, and real cascades propagate. We do not simulate any part of $T$.

The resulting corpus contains 27{,}243 verified transition samples spanning 64 worlds across 6 industries (financial services, government, healthcare, manufacturing, retail, technology) and 3 organizational sizes (small, midmarket, enterprise). We construct train/test splits at the world level, stratified by industry and organizational size, and reserve held-out industry--size combinations for evaluation. As a result, evaluation requires generalizing to unseen deployment regimes rather than interpolating among samples from the same worlds.

\paragraph{Benchmarking.}
We construct \emph{CascadeBench}, to evaluate transition prediction under controlled configuration shift. CascadeBench retains WoW's evaluation methodology: models predict field-level state changes from a proposed action, and predictions are compared against audit-log ground truth. However, CascadeBench is designed to isolate \emph{reasoning} over provided rules from confounding factors present in existing benchmarks, including parametric memorization, retrieval noise, and audit-log artifacts.
 
Concretely, CascadeBench differs from existing enterprise benchmarks along three axes. First, it is built on synthetic schemas that do not appear in real platform deployments, so models cannot rely on memorized table structures. Second, CascadeBench makes the relevant context available for each example---table schemas, business rules, and seed records---so we can control how much context the model receives. This enables both fully contextualized evaluation and context-limited settings that probe internalized knowledge or the effectiveness of runtime discovery. Third, audit-log ground truth is restricted to content fields, removing engine-internal metadata that does not reflect business logic, such as system identifiers, timestamps, and bookkeeping fields. Together, these choices let CascadeBench disentangle memorization, context discovery, and rule-based reasoning. We describe the construction pipeline in Appendix~\ref{app:wowpp}. We provide a comparison between CascadeBench and WoW in Appendix~\ref{app:wow_vs_cb}.

\section{Approaches}
\label{sec:approaches}

We compare three approaches to predict enterprise state transitions: prompting a frozen model, fine-tuning a learned world model, and using a discovery agent that inspects the current instance at inference time. All approaches take a current state $s_t$ and proposed action $a_t$ as input and output the same target representation: structured field-level diffs describing the predicted transition.

\subsection{Prompted Baseline}
\label{sec:prompted_baseline}

The prompted baseline uses a frozen language model to predict the effects of an action from the provided context alone. Given $s_t$ and $a_t$, the model outputs the expected state change as a structured set of field-level diffs. This baseline measures how well general-purpose models can infer transition behavior without fine-tuning or runtime access to instance-specific configuration. Depending on the evaluation setting, the prompt may include only the action and relevant state, or additional provided context such as schemas and rules.

\subsection{Learned Enterprise World Model}
\label{sec:trained_wm}

The learned enterprise world model predicts transitions by 
internalizing dynamics from supervised data. We fine-tune on 
$(s_t, a_t, s_{t+1})$ tuples collected from Enterprise Gym 
(\S\ref{sec:gym}), with the target being the minimal field-level 
diff between $s_t$ and $s_{t+1}$. This tests whether learned 
dynamics transfer to instances with different industries, 
organizational structures, and rule sets.






\subsection{Enterprise Discovery Agent}
\label{sec:discovery}

The enterprise discovery agent predicts the outcome of a proposed action without executing it and without updating model parameters. Unlike learned world models, it does not attempt to internalize environment dynamics. Instead, it queries the live instance configuration $c$ and reasons over the retrieved information to infer the effects of an action.

We model enterprise transitions as depending on instance-specific configuration:
\begin{equation}
s_{t+1} \sim P(s_{t+1} \mid s_t, a_t, c),
\end{equation}
where $c$ denotes the deployed configuration of the current instance. Since $c$ can be large, the discovery agent follows a retrieve-then-reason strategy. Given $s_t$ and $a_t$, it retrieves a task-relevant subset $\tilde{c} \subseteq c$ and predicts the next state as
\begin{equation}
\hat{s}_{t+1} = f_{\text{LLM}}\!\left(s_t,\; a_t,\; \tilde{c},\; \hat{s}_{1:t}\right),
\end{equation}
where $f_{\text{LLM}}$ is a frozen language model and $\hat{s}_{1:t}$ denotes prior predictions in a multi-step rollout.

Retrieval is adaptive: simple transitions may require little or no
additional context, while more complex cascades trigger targeted
queries for relevant rules, schemas, records, or SLA definitions. For multi-step rollouts, predictions are generated sequentially,
with each $\hat{s}_i$ appended to the context before predicting
$\hat{s}_{i+1}$, enabling the agent to reason about compounding
effects across the cascade chain (\S\ref{sec:depth_analysis}). Because $\tilde{c}$ is
retrieved at inference time rather than memorized during training,
the same agent transfers across tenants of the same enterprise
platform without modification. To isolate the contribution of
runtime discovery, the static context is matched to that of
prompted baselines; any improvement can therefore be attributed
to retrieval and reasoning over $\tilde{c}$. Implementation details
for the enterprise discovery agent can be found in Appendix~\ref{app:discovery_agent}.

\begin{table*}[t]
\caption{\textbf{Main transition-prediction results on CascadeBench and WoW.} We report IoU and table/field-level IoU (IoU(T+F)) with and without access to business rules (BR). Bold indicates the best score in each column for each model type.}
\centering
\label{tab:main_results}

\resizebox{\textwidth}{!}{%
\begin{tabular}{
>{\centering\arraybackslash}m{3.0cm} 
l 
cccccc}
 &  
& \multicolumn{4}{c}{\textbf{CascadeBench}} 
& \multicolumn{2}{c}{\textbf{WoW}} \\
\cmidrule(lr){3-6} \cmidrule(lr){7-8}
& 
& \multicolumn{2}{c}{\textbf{w/ BR}} 
& \multicolumn{2}{c}{\textbf{w/o BR}} 
& \multicolumn{2}{c}{\textbf{w/o BR}} \\[-0.5ex]
\cmidrule(lr){3-4} \cmidrule(lr){5-6} \cmidrule(lr){7-8}
\textbf{Type} & \textbf{Model} 
& \textbf{IoU} & \textbf{IoU(T+F)} 
& \textbf{IoU} & \textbf{IoU(T+F)} 
& \textbf{IoU} & \textbf{IoU(T+F)} \\
\midrule
\rowcolor{frontierbg}
& Sonnet 4.6~\citep{anthropic2026claudesonnet46}   & 38.15 & 57.67 & 10.30 & 10.53 & 38.65 & 41.54 \\
\rowcolor{frontierbg}
& Opus 4.6\citep{anthropic2026claudeopus46}     & 40.46 & 59.69 & \textbf{10.91} & \textbf{16.15} & \textbf{41.32} & \textbf{44.90} \\
\rowcolor{frontierbg}
& GPT-5\citep{openai2025gpt52systemcard}        & \textbf{41.78} & \textbf{61.50} & 9.82  & 13.25 & 29.34 & 32.32 \\
\rowcolor{frontierbg}
\multirow{-4}{=}{\centering \textbf{Frontier Models}}
& Gemini 3 Pro\citep{googledeepmind2025gemini3pro} & 41.36 & 59.53 & 10.38 & 13.92 & 35.59 & 39.06 \\

\midrule
\rowcolor{basebg}
& Qwen-3.5-27B\citep{qwen3.5} & \textbf{40.39} & \textbf{60.86} & 9.77 & 13.18 & 21.22 & 15.19 \\
\rowcolor{basebg}
& Qwen-3.6-27B\citep{qwen3.6-27b} & 38.98 & 60.06 & 7.13 & 13.15 & \textbf{23.17} & \textbf{27.02} \\
\rowcolor{basebg}
\multirow{-3}{=}{\centering \textbf{Base Models}}
& Gemma-4-31B\citep{google_gemma_model_card}  & 37.99 & 60.32 & \textbf{9.88} & \textbf{13.76} & 21.11 & 23.55 \\

\midrule
\rowcolor{finetunedbg}
& Qwen-3.5-27B-LoRA & \textbf{50.90} & \textbf{61.47} & 10.60 & 14.51 & 31.21 & 35.02 \\
\rowcolor{finetunedbg}
& Qwen-3.6-27B-LoRA & 40.47 & 60.93 & 9.74  & 13.40 & \textbf{32.22} & \textbf{36.88} \\
\rowcolor{finetunedbg}
\multirow{-3}{=}{\centering \textbf{Finetuned Models}}
& Gemma-4-31B-LoRA  & 41.33 & 52.62 & \textbf{12.19} & \textbf{16.41} & 31.73 & 36.07 \\
\bottomrule
\end{tabular}%
}
\vspace{-10pt}
\end{table*}

\section{Experiments}
\label{sec:experiments}

We organize the analysis as a three-rung ladder. Each rung is a declarative claim about where the dynamics come from at prediction time, with evidence drawn from Table~\ref{tab:main_results}.

\paragraph{Models.} We fine-tune Qwen-3.5-27B~\citep{qwen3.5}, Qwen-3.6-27B~\citep{qwen3.6-27b}, and Gemma-4-31B-it~\citep{google_gemma_model_card} with LoRA~\citep{hu2022lora} on the transition tuples from \S\ref{sec:gym}. The same models are evaluated zero-shot as prompted baselines, together with frontier models (Claude Sonnet~4.6~\citep{anthropic2026claudesonnet46}, Claude Opus~4.6~\citep{anthropic2026claudeopus46}, GPT-5~\citep{singh2025openai}, Gemini~3 Pro~\citep{googledeepmind2025gemini3pro}).

\paragraph{Metrics.} All methods take $(s_t, a_t)$ as input and predict 
field-level diffs, which we score against audit-log ground truth using 
two complementary IoU variants from~\citet{gupta2026world}. \textbf{IoU(T+F)} 
credits a prediction when it correctly identifies the affected 
\emph{(table, field)} pair, capturing whether the model has identified 
\emph{what changes} in the global state. Strict \textbf{IoU} additionally 
requires the predicted value to match, capturing \emph{how it changes}. 
We report both because identifying which elements of the global state 
will be impacted is itself a substantial part of the prediction problem 
in enterprise environments---a state diff over a database with thousands 
of fields requires the model to first localize the cascade footprint 
before reasoning about specific values.

\paragraph{Evaluation settings.} On \emph{CascadeBench} we run two settings: \textbf{w/ BR} supplies the relevant business rules in the prompt (an oracle for retrieval), and \textbf{w/o BR} removes them (testing what the model knows on its own). 
We also report results on the WoW benchmark, which runs the same prediction task on real ServiceNow instances with no business rules in the prompt.

\label{sec:rung1}

\paragraph{Rung 1: Prompting alone struggles when rules are hidden; SFT helps mainly without rule context.}
Table~\ref{tab:main_results} shows that when business rules are not provided, prompted models perform poorly on CascadeBench, with both frontier and base open-weight models in the 9--16 IoU(T+F) range. This is substantially lower than the 21--23 range observed for base models on WoW, suggesting that CascadeBench poses a harder transition-prediction problem with more hidden or cascading dynamics. SFT on the transition tuples from \S\ref{sec:gym} improves this no-BR setting, yielding modest gains on CascadeBench w/o BR (${\sim}2$--$3$ IoU points) and larger gains on WoW (${\sim}10$ points). With business rules in context, however, SFT is not uniformly beneficial: Qwen-3.5-27B reaches 50.9 IoU, above the 38--42 range of prompted models with BRs, but other models gain little or regress. Thus, SFT helps when rule context is missing, but does not consistently substitute for grounding in the active rules.

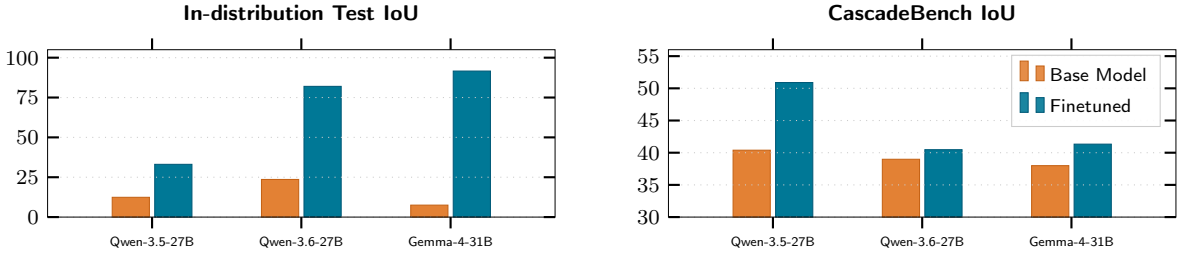
\begin{figure*}[h!]
\centering
\small
\vspace{-8pt}
\begin{tikzpicture}
\begin{axis}[
  name=ax1,
  ybar,
  ymin=0, ymax=105,
  width=0.5\textwidth, height=3.8cm,
  bar width=14pt,
  enlarge x limits=0.35,
  symbolic x coords={Qwen-3.5-27B, Qwen-3.6-27B, Gemma-4-31B},
  xtick=data,
  xticklabel style={font=\tiny\sffamily},
  ytick={0,25,50,75,100},
  yticklabel style={font=\footnotesize\sffamily},
  title style={font=\footnotesize\sffamily\bfseries, yshift=2pt},
  title={In-distribution Test IoU},
  axis line style={black, line width=0.3pt},
  tick style={black, thick},
  ymajorgrids=true,
  grid style={dotted, gray!40},
  axis on top,
]
\addplot[fill={rgb,255:red,227;green,129;blue,52}, draw={rgb,255:red,197;green,99;blue,22}] coordinates {
  (Qwen-3.5-27B, 12.4)
  (Qwen-3.6-27B, 23.6)
  (Gemma-4-31B,  7.50)
};
\addplot[fill={rgb,255:red,0;green,120;blue,150}, draw={rgb,255:red,0;green,90;blue,120}] coordinates {
  (Qwen-3.5-27B, 33.09)
  (Qwen-3.6-27B, 82.00)
  (Gemma-4-31B,  91.60)
};
\end{axis}

\begin{axis}[
  at={(ax1.north east)},
  anchor=north west,
  xshift=1.5cm,
  ybar,
  ymin=30, ymax=56,
  width=0.5\textwidth, height=3.8cm,
  bar width=14pt,
  enlarge x limits=0.35,
  symbolic x coords={Qwen-3.5-27B, Qwen-3.6-27B, Gemma-4-31B},
  xtick=data,
  xticklabel style={font=\tiny\sffamily},
  ytick={30,35,40,45,50,55},
  yticklabel style={font=\footnotesize\sffamily},
  title style={font=\footnotesize\sffamily\bfseries, yshift=2pt},
  title={CascadeBench IoU},
  axis line style={black, line width=0.3pt},
  tick style={black, thick},
  ymajorgrids=true,
  grid style={dotted, gray!40},
  axis on top,
  legend style={
    at={(0.97, 0.97)},
    anchor=north east,
    font=\scriptsize\sffamily,
    draw=gray!40,
    fill=white,
    fill opacity=0.9,
    text opacity=1,
    row sep=1pt,
    legend cell align=left,
  },
]
\addplot[fill={rgb,255:red,227;green,129;blue,52}, draw={rgb,255:red,197;green,99;blue,22}] coordinates {
  (Qwen-3.5-27B, 40.39)
  (Qwen-3.6-27B, 38.98)
  (Gemma-4-31B,  37.99)
};
\addplot[fill={rgb,255:red,0;green,120;blue,150}, draw={rgb,255:red,0;green,90;blue,120}] coordinates {
  (Qwen-3.5-27B, 50.90)
  (Qwen-3.6-27B, 40.47)
  (Gemma-4-31B,  41.33)
};
\legend{Base Model, Finetuned}\end{axis}
\end{tikzpicture}
\vspace{-8pt}
\caption{In-distribution Test IoU \& CascadeBench IoU (out-of-distribution) comparing base models and finetuned counterparts. All settings have access to business rules.}
\vspace{-8pt}
\label{fig:iou-bars}
\end{figure*}

\label{sec:rung2}

\paragraph{Rung 2: SFT is strong in distribution but degrades under shift.} Figure~\ref{fig:iou-bars} shows that fine-tuning can strongly internalize the training dynamics: in-distribution test IoU rises to 91.6 for Gemma-4-31B and 82.0 for Qwen-3.6-27B, far above their base counterparts. However, this advantage largely collapses on CascadeBench, where models face synthetic schemas and configurations not seen during training: both models fall to roughly 40--41 IoU. Fine-tuned models remain stronger than prompted baselines, but most of their in-distribution edge is lost under shift. This suggests that SFT learns useful transition patterns, but also binds them to the training distribution; internalization alone is therefore insufficient for robust cross-instance prediction.

\begin{table*}[t]
\caption{\textbf{Discovery agent vs.\ oracle and prompted models on CascadeBench.} 
\emph{Oracle} provides business rules in context (reasoning ceiling). 
\emph{DA} starts without rules recovers them via retrieval at 
inference. \emph{Prompted} is the no-context floor.}
\centering
\small
\setlength{\tabcolsep}{6pt}
\resizebox{\textwidth}{!}{%
\begin{tabular}{
>{\centering\arraybackslash}m{3.0cm} 
l 
cccccc}
\toprule
&& \textbf{Oracle} & \textbf{DA}  & \textbf{Prompted} \\
\textbf{Type} & \textbf{Model} & \textbf{w/ BR}  & \textbf{w/o BR} & \textbf{w/o BR} \\
\midrule

\rowcolor{frontierbg}
& GPT-5~\citep{openai2025gpt52systemcard}             & 41.78 & 32.1  & 9.82  \\
\rowcolor{frontierbg}
& Claude Opus 4.6~\citep{anthropic2026claudeopus46}   & 40.46 & 32.0  & 10.91 \\
\rowcolor{frontierbg}
& Claude Sonnet 4.6~\citep{anthropic2026claudesonnet46} & 38.15 & 29.7  & 10.30 \\
\rowcolor{frontierbg}
\multirow{-4}{=}{\centering \textbf{Frontier Models}} & Gemini 3 Pro~\citep{googledeepmind2025gemini3pro}     & 41.36 & 31.2  & 10.38 \\

\midrule

\rowcolor{finetunedbg}
& Qwen-3.6-27B-LoRA~\citep{qwen3.6-27b} & 40.47 & \textbf{28.8} & 9.74  \\
\rowcolor{finetunedbg}
& Qwen-3.5-27B-LoRA~\citep{qwen3.5} & 50.90 & \textbf{21.5} & 10.60 \\
\rowcolor{finetunedbg}
\multirow{-3}{=}{\centering \textbf{Finetuned Models}}
& Gemma-4-31B-LoRA~\citep{google_gemma_model_card}  & 41.33 & 12.5 & 12.19 \\

\bottomrule
\end{tabular}
}
\label{tab:da_cascadebench}
\end{table*}

\label{sec:rung3}

\begin{figure}[t]
\vspace{-8pt}
    \centering
    \includegraphics[width=\linewidth]{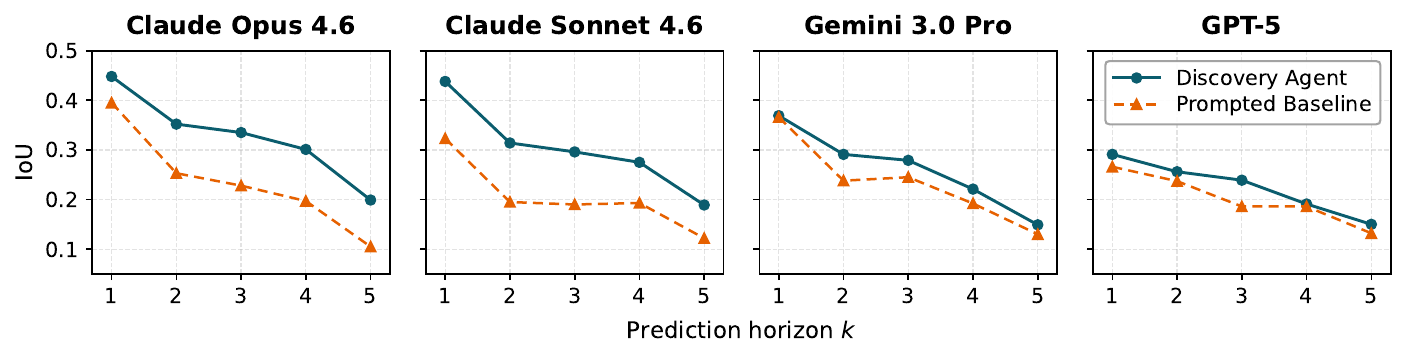}
    \vspace{-20pt}
    \caption{\textbf{Runtime discovery improves multi-step state prediction.}
IoU across prediction horizons $k=1,\ldots,5$ for four backbone models on World of Workflows.
Performance degrades as the rollout horizon increases, but the Discovery Agent remains consistently above the prompted baseline, indicating that retrieving transition logic at inference time helps reduce compounding prediction errors.}
    \label{fig:state_prediction}
    \vspace{-12pt}
\end{figure}

\paragraph{Rung 3: Runtime discovery recovers cross-instance accuracy.}
If neither the prompt nor the weights solve the problem on their own, the remaining option is to recover the rules from the live instance at inference time. The discovery agent (\S\ref{sec:discovery}) uses the same static context as the prompted baseline, but can additionally query the live instance for rules, schemas, and records before predicting. Figure~\ref{fig:state_prediction} shows state-prediction IoU on WoW across rollout horizons $k{=}1,\ldots,5$. Discovery improves over the matched prompted baseline for every model and horizon we evaluate, including at $k{=}1$, where Opus~4.6 rises from 0.40 to 0.45 and Sonnet~4.6 from 0.32 to 0.44. The margin varies across settings: discovery and prompting are sometimes close, but discovery also yields substantial gains of up to roughly 0.10 IoU. Importantly, the advantage remains visible across rollout depths despite compounding errors. This suggests that querying the live instance provides complementary signal beyond the static prompt, improving cross-instance prediction without training on the target instance.

\begin{figure}[t]
    \centering
    \includegraphics[width=\linewidth]{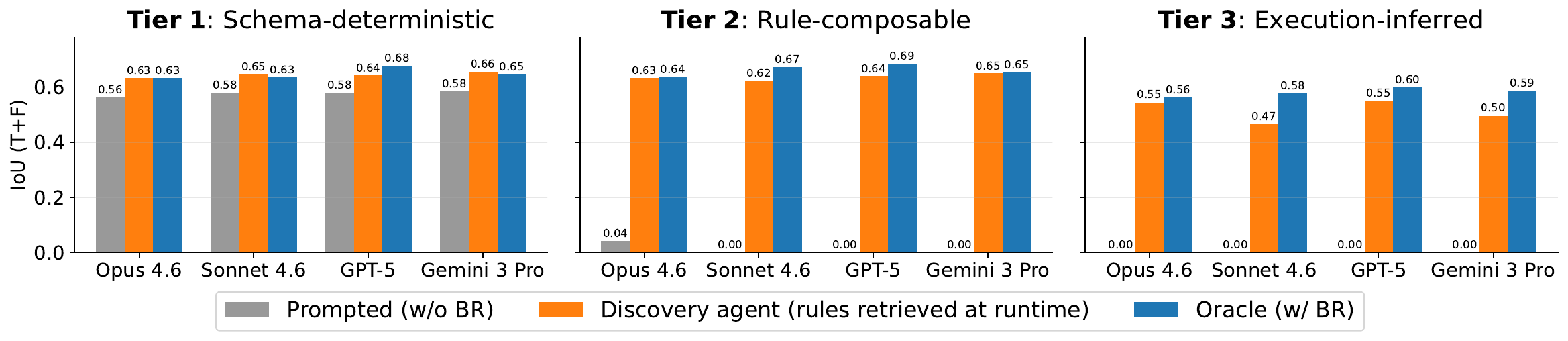}
    \vspace{-15pt}
    \caption{\textbf{Tier-stratified transition prediction on CascadeBench.} Prompted models are often sufficient for simple schema-determined effects, but fail on harder tiers where hidden workflows, rule cascades, and execution-dependent effects shape the next state. Runtime discovery recovers much of the information available to the oracle rule-in-context setup.}
    \label{fig:tier_analysis}
    \vspace{-10pt}
\end{figure}

\paragraph{Retrieval beats internalization on the same model.} Table~\ref{tab:da_cascadebench} compares three ways of accessing transition logic on CascadeBench: providing business rules in context, retrieving them at inference time, or relying on the prompt/weights without retrieval. For the non-finetuned frontier models, the static no-BR baseline is consistently low around 10 IoU, while the discovery agent recovers a large fraction of the oracle signal, reaching the mid-20s to low-30s without any training on the target instance. This shows that the benefit of discovery is not specific to fine-tuned models. On the LoRA models, the same-model comparison is more nuanced: retrieval substantially outperforms internalization for the Qwen models, while Gemma is roughly tied and remains far below the oracle setting. Overall, runtime retrieval is a more reliable source of cross-instance signal than static prompting or weights alone, but the remaining gap to the oracle indicates that retrieval and rule composition are still imperfect. This motivates training discovery agents to retrieve and compose the active rules more effectively.

\paragraph{Discovery helps most when transition logic exceeds the schema.}
Figure~\ref{fig:tier_analysis} stratifies CascadeBench by transition complexity for proprietary models, following \S\ref{sec:dynamics}. Prompting without rules handles Tier~1 schema effects reasonably well, reaching roughly 0.56--0.58 IoU, but nearly collapses on Tier~2 cascades and Tier~3 conflicts because rule context is missing. Runtime discovery recovers most of this gap: across Claude Opus~4.6, Claude Sonnet~4.6, GPT-5, and Gemini~3 Pro, it stays near the oracle on Tier~1 and Tier~2 while outperforming the prompted baseline. The main remaining gap is Tier~3, where outcomes depend on execution semantics not fully exposed in configuration. Thus, discovery is most valuable where static prompting fails: hidden rules and cascades, not schema-only effects.





\section{Discussion}
\label{sec:disc}

\paragraph{Effect of Business Rules.} \label{sec:br_effect}
Across the model classes in Table~\ref{tab:main_results}, removing business rules from the prompt produces a uniform collapse on CascadeBench. With BR, every class --- frontier, base, and SFT --- sits in a 38--51 IoU band; without BR, the same models fall to 7--12. The drop is consistent across model sizes and families. This demonstrates that \emph{business rules carry the dynamics that CascadeBench measures}: the benchmark probes rule-grounded reasoning rather than what models already know from pretraining. Furthermore, since rules cannot be replaced by pretraining or fine-tuning, supplying them at inference time in the prompt or via runtime retrieval is the deciding factor for prediction accuracy.

\paragraph{Depth Analysis.}
\label{sec:depth_analysis}
Figure~\ref{fig:state_prediction} shows that Discovery Agents outperform matched prompted baselines across rollout horizons. Performance generally decreases as $k$ grows for both methods, as longer horizons require predicting deeper cascades and create more opportunities for error accumulation. Despite this increasing difficulty, discovery remains consistently beneficial from $k{=}1$ through $k{=}5$. Across all matched models and rollout horizons, the Discovery Agent improves over the prompted baseline, with the size of the gain varying by model and depth.

The likely mechanism is repeated grounding. Both methods condition later predictions on earlier predicted diffs, so neither is immune to compounding errors. However, the prompted baseline relies primarily on its initial context and prior outputs, while the Discovery Agent can re-query the live instance for relevant records, active business rules, and reference identifiers at each step. This refreshes the model's view of the deployed configuration rather than relying only on its evolving prediction state. Thus, the same backbone model produces stronger long-horizon predictions when placed inside a discovery loop, showing that runtime retrieval improves robustness in multi-step cascade prediction.

\section{Conclusion}
\label{sec:conclusion}

This paper studies transition prediction in enterprise environments, where dynamics are shaped by tenant-specific configurations rather than fixed rules inferred only from experience. We find that offline-trained world models perform well in-distribution but degrade on held-out configurations. Discovery agents, which retrieve relevant rules at inference time, remain more robust under shift and avoid some of the error compounding observed in purely internalized models.

Discovery agents are not a replacement for learned world models. Instead, our results suggest that when transition logic is readable from the live system, agents should not rely solely on internalized dynamics. The next step is to combine learned priors with runtime retrieval and reasoning: training agents that learn when, what, and how to retrieve. We discuss the scope and assumptions of this conclusion in Section~\ref{sec:limitations}.

\section{Limitations}
\label{sec:limitations}

The discovery agent assumes business rules are readable on the 
live instance; access controls degenerate it to the prompted 
baseline. DA performance also depends on tool-use capability: on 
open-weight models in the 27--31B range, the retrieval loop is 
unreliable enough that LoRA finetuning wins in some conditions, 
so the choice between training and discovery is deployment-dependent. 
Our evaluation is single-platform (ServiceNow) and our 
quantitative results focus on Tier~1 and Tier~2 transitions; 
Tier 3 results are reported in Appendix \ref{app:tier_stratified}; Tier 3 stratification is limited to multi-rule conflicts detectable from the audit log, and broader execution-order dynamics remain out of scope. The DA-vs-trained comparison rests on a small 
set of open-weight models where LoRA finetuning is feasible. 
Appendix~\ref{app:limitations} expands on each.

\bibliography{references}
\bibliographystyle{servicenow}

\appendix
\newpage

\section*{\LARGE Appendix}
\vspace{8pt}

\noindent\textbf{\Large Table of Contents}
\begin{flushright}
    \textbf{Page}
\end{flushright}

\noindent
\renewcommand{\arraystretch}{1.2}
\begin{tabularx}{\linewidth}{Xr}
    \textbf{A. Limitations} \dotfill & \pageref{app:limitations} \\
    \textbf{B. Discovery Agent Scores} \dotfill & \pageref{app:da_scores} \\
    \textbf{C. Tier-Stratified Results} \dotfill & \pageref{app:tier_stratified} \\
    \textbf{D. CascadeBench: Construction Pipeline} \dotfill & \pageref{app:wowpp} \\
    \textbf{E. Enterprise Gym: World Construction Details} \dotfill & \pageref{app:gym} \\
    \textbf{F. WoW vs CascadeBench} \dotfill & \pageref{app:wow_vs_cb} \\
    \textbf{G. Discovery Agent Implementation} \dotfill & \pageref{app:discovery_agent} \\
    \textbf{H. CascadeBench: Failure Mode Analysis} \dotfill & \pageref{app:cascade_analysis} \\
    \hspace{2em} H.1 Aggregate Failure Analysis \dotfill & \pageref{app:failure_analysis} \\
    \textbf{I. Fine-tuning details} \dotfill & \pageref{app:finetuning_details} \\
    \textbf{J. Glossary} \dotfill & \pageref{app:glossary} \\
\end{tabularx}

\newpage
\section{Limitations}
\label{app:limitations}

\paragraph{Inspectability assumption.} The discovery agent assumes the 
relevant business rules and supporting tables are readable on the live 
instance. Production deployments often impose access controls, in which case runtime discovery degenerates to the prompted 
baseline.

\paragraph{Tool-use capability bounds discovery.} DA performance depends 
on the model's ability to issue and reason over tool calls. On 
open-weight models in the 27--31B range, the retrieval loop is 
unreliable enough that DA underperforms LoRA-finetuned variants on the 
same model in some conditions (Gemma-4-31B-LoRA, 
Table~\ref{tab:da_cascadebench}). The right choice between training and 
discovery is therefore deployment-dependent: frontier APIs favor 
discovery, constrained open-weight deployments favor finetuning.

\paragraph{Single-platform evaluation.} Our experiments evaluate on 
ServiceNow. Other enterprise platforms have 
different rule formalisms, cascade semantics, and inspectability 
guarantees. We expect the inversion argument to transfer 
(configurability is a property of all major enterprise platforms) but 
do not demonstrate it directly.

\paragraph{Tier 3 dynamics.} Tier-stratified results, including a Tier 3
stratum, are reported in Appendix~\ref{app:tier_stratified}. Our Tier 3
operationalization is restricted to multi-rule conflicts where two or
more rules write distinct values to the same field, which are
detectable from the audit log. Broader execution-inferred dynamics,
such as async/sync interleaving, race conditions across parallel rule
firings, and platform-internal scheduling behaviors, are present in
the Enterprise Gym corpus by construction but are not separately
stratified, since attributing them at scale requires resolving
execution-order semantics that the audit log alone does not expose.
Extending the Tier 3 stratum to cover these cases is left to follow-up
work.

\paragraph{Same-model comparison scope.} The DA-vs-trained comparison 
in Table~\ref{tab:da_cascadebench} is restricted to models where LoRA 
finetuning is feasible (Qwen-3.5/3.6-27B, Gemma-4-31B). The finding 
that retrieval beats internalization on a fixed model therefore rests 
on a small set of open-weight models.
\section{Discovery Agent Scores}
\label{app:da_scores}

\begin{table*}[t]
\caption{Cascade prediction performance on WoW across prediction
horizons $k$. Discovery agents (DA) on capable models outperform
prompted baselines at every horizon; on weaker open-weight models,
DA performance is bounded by the model's tool-use capability.}
\centering
\small
\setlength{\tabcolsep}{4pt}
\begin{tabular}{ll ccccc}
\toprule
\textbf{Method} & \textbf{Models} & \textbf{$k=1$} & \textbf{$k=2$} & \textbf{$k=3$} & \textbf{$k=4$} & \textbf{$k=5$} \\
\midrule
\multirow{6}{*}{\textit{Prompted}}
  & GPT-5 Mini       & 0.190 & 0.142 & 0.139 & 0.094 & 0.075 \\
  & GPT-5            & 0.266 & 0.237 & 0.186 & 0.186 & 0.132 \\
  & Gemini 3 Pro     & 0.366 & 0.238 & 0.245 & 0.192 & 0.130 \\
  & Sonnet 4.5       & 0.358 & 0.211 & 0.238 & 0.199 & 0.134 \\
  & Sonnet 4.6       & 0.323 & 0.195 & 0.190 & 0.193 & 0.122 \\
  & Opus 4.6         & 0.395 & 0.253 & 0.228 & 0.197 & 0.105 \\
\midrule
\multirow{8}{*}{\textit{Discovery (ours)}}
  & Gemma-4-31B      & 0.214 & 0.201 & 0.154 & 0.134 & 0.095 \\
  & Qwen-3.6-27B     & 0.227 & 0.210 & 0.175 & 0.189 & 0.109 \\
  & GPT-5 Mini       & 0.258 & 0.222 & 0.200 & 0.164 & 0.101 \\
  & GPT-5            & 0.291 & 0.256 & 0.239 & 0.191 & 0.150 \\
  & Gemini 3 Pro     & 0.369 & 0.291 & 0.279 & 0.221 & 0.149 \\
  & Sonnet 4.5       & 0.406 & 0.326 & 0.329 & 0.274 & 0.186 \\
  & Sonnet 4.6       & 0.438 & 0.314 & 0.296 & 0.275 & 0.189 \\
  & \textbf{Opus 4.6}        & \textbf{0.448} & \textbf{0.352} & \textbf{0.335} & \textbf{0.301} & \textbf{0.199} \\
\bottomrule
\end{tabular}
\label{tab:wow_results_horizons}
\end{table*}

Table~\ref{tab:wow_results_horizons} reports the discovery agent's 
performance on WoW across all evaluated models and prediction 
horizons $k=1, \ldots, 5$, alongside the corresponding prompted 
baselines on the same models. The DA improves over the matched 
prompted baseline at every horizon on every model where both are 
evaluated. The improvement is largest at intermediate horizons 
($k=2$ to $k=4$), where small static-context errors begin to 
compound but the discovery agent's retrieval still tracks the 
evolving state.

\section{Tier-Stratified Results}
\label{app:tier_stratified}

We stratify CascadeBench results along the three tiers introduced
in \S\ref{sec:dynamics} (Table~\ref{tab:complexity_tiers}). The
tier definitions in \S\ref{sec:dynamics} are conceptual; here we
operationalize them for measurement: T1 covers schema-determined
effects on the action's own table, which exercise the data-dictionary
defaults and constraints described in Tier~1; T2 covers cross-table
cascades requiring at least one business rule to fire, instantiating
the rule-composable cascades of Tier~2; T3 covers multi-rule
conflicts where two or more rules write distinct values to the same
field, an audit-log-detectable instance of the execution-inferred
dynamics described in Tier~3. T1 and T2 are scored under IoU(T+F);
T3 is scored under Strict IoU on
$(\text{table}, \text{field}, \text{value})$, since the defining
question for T3 is which conflicting value is realized. System
metadata fields, datetime values, and reference identifiers are
excluded symmetrically from ground truth and predictions, as they
reflect platform internals rather than the business logic under
evaluation. Bootstrap 95\% confidence intervals
($n = 2{,}000$ resamples) are reported in the supplementary
materials.

\begin{table}[h]
\centering
\small
\caption{Tier-stratified IoU on CascadeBench ($k = 1$, 37
trajectories; T1: 177 keys, T2: 424 keys, T3: 138 keys).
\emph{Direct}: prompted without retrieval and without rules.
\emph{Discovery Agent}: rules retrieved at inference.
\emph{Oracle}: rules in prompt.}
\label{tab:tier_stratified}
\begin{tabular}{l c c c c}
\toprule
Model / Setting & ALL & T1 & T2 & T3 \\
\midrule
\multicolumn{5}{l}{\textit{Direct (no rules, no retrieval)}} \\
Opus 4.6            & 0.175 & 0.562 & 0.043 & 0.000 \\
Sonnet 4.6          & 0.152 & 0.580 & 0.000 & 0.000 \\
GPT-5               & 0.152 & 0.580 & 0.000 & 0.000 \\
Gemini 3 Pro        & 0.153 & 0.584 & 0.000 & 0.000 \\
Qwen-3.5-27B        & 0.151 & 0.578 & 0.000 & 0.000 \\
Qwen-3.6-27B        & 0.151 & 0.578 & 0.000 & 0.000 \\
Gemma-4-31B         & 0.157 & 0.601 & 0.000 & 0.000 \\
\midrule
\multicolumn{5}{l}{\textit{Discovery Agent (rules retrieved at inference)}} \\
Opus 4.6            & 0.638 & 0.633 & 0.633 & 0.545 \\
Sonnet 4.6          & 0.638 & 0.646 & 0.624 & 0.466 \\
GPT-5               & 0.646 & 0.642 & 0.639 & 0.551 \\
Gemini 3 Pro        & 0.656 & 0.657 & 0.650 & 0.497 \\
\midrule
\multicolumn{5}{l}{\textit{Oracle (rules in prompt)}} \\
Opus 4.6            & 0.643 & 0.632 & 0.637 & 0.564 \\
Sonnet 4.6          & 0.669 & 0.634 & 0.674 & 0.578 \\
GPT-5               & 0.691 & 0.679 & 0.686 & 0.599 \\
Gemini 3 Pro        & 0.659 & 0.647 & 0.654 & 0.588 \\
Qwen-3.5-27B        & 0.669 & 0.646 & 0.669 & 0.554 \\
Qwen-3.6-27B        & 0.660 & 0.635 & 0.661 & 0.572 \\
Gemma-4-31B         & 0.694 & 0.663 & 0.695 & 0.549 \\
\bottomrule
\end{tabular}
\end{table}

\paragraph{T1 is predictable from the action and schema alone.}
Direct IoU on T1 is consistent across all eight models at
$0.56$--$0.60$. Schema following is sufficient for action-table
effects, and rule retrieval is not required. T2 and T3 drop to
$0.00$ uniformly under Direct prompting, identifying business
rules as the load-bearing signal in CascadeBench.

\paragraph{Discovery exhibits graded degradation across tiers.}
Under the Discovery Agent, mean IoU decreases monotonically from
T1 ($0.648$) through T2 ($0.635$) to T3 ($0.524$). The Oracle
condition is flatter (T1: $0.634$, T2: $0.635$, T3: $0.569$),
indicating that the T1$\to$T2 gap under DA reflects retrieval
overhead rather than reasoning difficulty.

\paragraph{Discovery reaches Oracle parity on T1 and T2; T3 bounds
both.} Across the five frontier models evaluated under both
conditions, mean DA $-$ Oracle deltas are $+0.014$ on T1,
$+0.001$ on T2, and $-0.046$ on T3. The Discovery Agent matches
the rule-oracle on the first two tiers without ground-truth rule
pre-loading. Both methods plateau on T3, where outcomes depend
on execution-order resolution that is not exposed in the
configuration either method reads. This is the empirical
realization of the boundary anticipated in
\S\ref{sec:dynamics}: dynamics determined by execution semantics
cannot be recovered from configuration alone.

Open-weight checkpoints reach the Oracle ceiling when rules are
supplied in prompt (Qwen-3.5-27B, Qwen-3.6-27B, and Gemma-4-31B
all reach ALL IoU $\geq 0.66$), indicating the determining
factor in this regime is rule content rather than model scale.
Whether a discovery loop on these checkpoints can reach the
same ceiling depends on tool-use capability, discussed in
\S\ref{app:limitations}.
\begin{table}[t]
  \centering
  \small
  \caption{\textbf{Complexity tiers of transition dynamics in enterprise systems.} Transitions range from fully determined by schema (Tier 1), to requiring composition of explicit rules (Tier 2), to behaviors that can only be inferred through execution (Tier 3).}
  \begin{tabular}{p{2.8cm} p{10cm}}
    \toprule
    \textbf{Tier} & \textbf{Description} \\
    \midrule

    \textbf{Tier 1} \newline \textit{Schema-Deterministic}
    & Transitions fully determined by the data dictionary (field types, defaults, constraints, choice lists); no rule inspection required. 
    
    \textbf{Example:} Creating a user sets: \newline\texttt{active=true}, \texttt{notification=2}, \texttt{locked\_out=false}. \\[0.5em]
    \addlinespace[0.35em]

    \textbf{Tier 2} \newline \textit{Rule-Composable}
    & Transitions determined by composing business rules; predictable given full rule knowledge, but requiring tracing of cascading execution. 
    
    \textbf{Example:} Setting priority to P1 triggers auto-assignment, starts an SLA timer, sends a notification, and creates an escalation record. \\[0.5em]
    \addlinespace[0.35em]

    \textbf{Tier 3} \newline \textit{Execution-Inferred}
    & Transitions depend on undocumented engine behavior, timing, or emergent interactions not recoverable from configuration alone. 
    
    \textbf{Example:} Race conditions between synchronous and asynchronous rules produce outcomes only observable through execution. \\

    \bottomrule
  \end{tabular}
    \label{tab:complexity_tiers}
\end{table}
\section{CascadeBench: Construction Pipeline}
\label{app:wowpp}
 
CascadeBench is generated by a three-stage pipeline that runs against a live ServiceNow instance: schema generation, business rule cascade construction, and cascade execution with audit capture. Each stage combines LLM-driven design with programmatic validation and live execution, and only fully validated, executable examples are included in the benchmark.
 
\paragraph{Schema generation.} An LLM proposes a synthetic enterprise domain (e.g., a vendor procurement workflow) and a corresponding schema: up to 10 tables with custom \texttt{u\_}-prefixed fields, choice values, and foreign-key relationships. Reserved namespace prefixes used by ServiceNow's product modules (\texttt{itam}, \texttt{cmdb}, \texttt{itsm}, \texttt{hr}, and others) are excluded so the schemas cannot overlap with platform tables seen during pretraining. The schema and accompanying seed records pass through deterministic structural validation (referential integrity, type conformance, graph connectivity, and naming constraints). Only schemas that satisfy all programmatic checks are deployed to the instance with auditing enabled and consistent seed data.
 
\paragraph{Business rule cascade construction.} Each example contains a cascade of 3--7 business rules that fire from a single triggering action. We support three cascade topologies: \emph{flat} (all rules fire from the same action on the primary table), \emph{linear} (each rule fires on a table written by a prior rule), and \emph{complete graph} (rules may fire on any table any prior rule has written to). Cascades are generated using weighted sampling over rule attributes (trigger type, conditions, fields updated, and tables impacted) to match realistic distributions. For each rule slot, an LLM designs the rule and its script. The script is statically analyzed to extract write operations, which serve as the source of truth for validation. Each rule is validated against 14 deterministic checks covering schema correctness, cascade integrity (including cycle detection), filter validity, and script safety. Failed rules are iteratively repaired; only rules that pass all checks and are verified through execution are included in the cascade.
 
\paragraph{Cascade execution and audit capture.}
Once the cascade is constructed, the pipeline deploys the rules to the instance, inserts any supporting records needed by the rule logic, and executes the triggering action. The platform's built-in audit log captures every field-level change that occurs, while a separate custom log traces each change back to the specific rule that caused it. Together, these two logs establish ground truth: one records \emph{what} changed, the other records \emph{why}. Before inclusion in the benchmark, internal metadata fields are filtered out, retaining only semantically meaningful content changes. After each cascade is captured, the instance is reset to its original seed state so that every example starts from identical initial conditions, preventing state leakage across examples.
\section{Enterprise Gym: World Construction Details}
\label{app:gym}

This appendix describes the construction pipeline behind each world $W = (E, T)$ in the Enterprise Gym (\ref{sec:gym}).

\paragraph{Dependency-ordered pipeline.} World construction follows a seven-stage pipeline where each stage depends on outputs from earlier stages, mirroring how real enterprise environments are built. Organizational structure (groups, users, roles) is generated first, followed by configuration database topology (services, infrastructure items, dependencies), then process design (per-domain state machines, routing, escalation), then business rules, access policies, SLA definitions, and finally field-level constraints. The ordering guarantees internal consistency: every entity referenced by a business rule has been materialized in an earlier stage. Rules that reference non-existent groups or infrastructure items produce deployment failures, not training data.

\paragraph{Variety engine.} Each archetype is seeded with a distinct organizational personality along multiple axes: automation philosophy (heavy vs. light reliance on rules), technical debt posture (clean vs. accumulated legacy automation), and domain completeness (which operational domains are fully built out vs. minimally configured). These dimensions ensure that two worlds in the same industry and company-size category still produce structurally distinguishable transition functions $T$, preventing the failure mode where a model trained on superficially diverse environments has effectively seen only one underlying transition function.

\paragraph{Conflict injection.} Rule conflicts are modeled at three levels: the pattern catalog encodes which rule types interfere with one another, variety profiles set per-archetype conflict density, and the generation stage produces specific conflicting rule pairs with explicit order-dependent behavior. This produces Tier~3 dynamics in the training data, transitions whose outcome depends on platform-internal execution ordering rather than on any single configuration artifact.

\paragraph{State-space augmentation.} From approximately 27{,}000 base scenarios generated by an LLM, a programmatic augmentation step produces ${\sim}$802{,}000 total initial states by swapping assignment groups, urgency/impact combinations, caller identities, and infrastructure references drawn from each archetype's validated entity pools. Seven quality guardrails ensure augmented scenarios remain internally consistent: pool validity (every substituted entity exists in the archetype), schema consistency, priority matrix coherence (urgency-impact-priority combinations match the archetype's priority calculator), referential integrity, business rule stripping during bulk insertion (to prevent rule-induced state changes from contaminating initial conditions), deduplication, and a final validation pass.

\section{WoW vs CascadeBench}
\label{app:wow_vs_cb}

CascadeBench provides every model with the full set of relevant
business rules and supporting context, simulating an oracle in which
retrieval is perfect. A model's CascadeBench score therefore represents
the upper bound on what a discovery agent built around that model could
achieve. WoW evaluates the same prediction task on real ServiceNow
instances with no provided context, so the gap to CascadeBench measures
how much real-world performance is left on the table by imperfect
grounding---precisely what discovery agents are designed to recover.

Table~\ref{tab:oracle_gap} shows that this gap is large and
model-dependent. Claude Sonnet 4.6 and Opus 4.6 close it almost
completely (gaps of 0.5 and 0.9 points), indicating their WoW failures
are bounded by reasoning capacity rather than missing context. GPT-5
and Qwen models show 12--19 point gaps, suggesting substantial
headroom that runtime retrieval should close. The discovery agent's
target is the oracle score; the gap to it is its remaining work.
\begin{table}[t]
\centering
\caption{CascadeBench (oracle) vs. WoW (no context) IoU. The gap is the headroom that runtime retrieval can recover.}
\label{tab:oracle_gap}
\small
\begin{tabular}{lccc}
\toprule
\textbf{Model} & \textbf{CascadeBench} & \textbf{WoW} & \textbf{Gap} \\
\midrule
Sonnet 4.6     & 38.15 & 38.65 & $+0.5$  \\
Opus 4.6       & 40.46 & 41.32 & $+0.9$  \\
Gemini 3 Pro   & 41.36 & 35.59 & $-5.8$  \\
Gemma-4-31B-FT & 41.32 & 31.73 & $-9.6$  \\
GPT-5          & 41.78 & 29.34 & $-12.4$ \\
GPT-5.2        & 38.40 & 23.97 & $-14.4$ \\
Qwen-3.5-27B   & 40.39 & 21.22 & $-19.2$ \\
Qwen-3.5-FT    & 50.90 & 31.21 & $-19.7$ \\
\bottomrule
\end{tabular}
\end{table}
\section{Discovery Agent Implementation}
\label{app:discovery_agent}

We implement the discovery agent as a ReAct-style~\cite{yao2022react} agent that predicts enterprise state transitions by inspecting the live system configuration at inference time. Unlike the learned world model, which relies on parameters learned from offline transition data, the discovery agent is given a proposed action and can query the active ServiceNow instance to recover the rules and records needed to predict its effects.

\paragraph{Input preparation.}
Each prediction begins with an initialization step that loads static context and validates the required state fields. The input state is normalized into JSON-serializable strings to ensure that records, dictionaries, and nested structures are represented consistently across examples. If a step index is not provided, we assign a default value so that multi-step trajectories and single-step examples share the same interface. The resulting prompt contains the table schemas, tool specification, previous record states, and the action whose effects must be predicted.

\paragraph{Agent setup.}
The discovery agent is composed as a SyGra~\citep{pradhan2025sygra} graph---an open-source framework that orchestrates LLM workflows---and instantiated as a ReAct agent with a single tool, \texttt{snow\_query}. The system prompt frames the model as an ITSM domain expert and instructs it to reason about how ServiceNow configuration artifacts determine state transitions. The agent is allowed up to 15 recursive tool calls. This budget is intended to support multi-hop discovery, where predicting a transition may require inspecting business rules, the current record state, choice lists, and SLA definitions before producing a final answer.

\paragraph{Runtime discovery loop.}
During inference, the agent alternates between reasoning and calls to \texttt{snow\_query}. The tool provides a uniform interface for querying ServiceNow tables relevant to the prediction task. In practice, the agent most commonly queries four classes of information:
\begin{enumerate}
    \item \textbf{Business rules} from \texttt{sys\_script}, which specify server-side transition logic triggered by record inserts or updates.
    \item \textbf{Current record state} from the target task table, which grounds the prediction in the active values of the affected record.
    \item \textbf{Choice values} from \texttt{sys\_choice}, which define valid categorical values and help interpret field-level updates.
    \item \textbf{SLA definitions} from \texttt{contract\_sla}, which capture additional transition logic related to task timing, priority, and service-level behavior.
\end{enumerate}

The agent uses these queries to identify which rules may fire for the proposed action, determine whether their conditions are satisfied, and infer the resulting field-level updates. The final agent response is required to contain a JSON object with an \texttt{audits} field, where each audit entry represents a predicted field-level change.

\paragraph{Output format.}
The agent predicts transitions in the same field-level diff format used by the benchmark. Each predicted audit record contains the affected table, field name, old value, and new value. This shared output format makes the discovery agent directly comparable to learned world models and prompted baselines. A typical output has the following structure:
\begin{verbatim}
{
  "audits": [
    {
      "tablename": "...",
      "fieldname": "...",
      "oldvalue": "...",
      "newvalue": "..."
    }
  ]
}
\end{verbatim}

\paragraph{Post-processing.}
Because the final response is produced by a language model, we apply a deterministic post-processing step before scoring. The \texttt{DiscoveryPostProcessor} first attempts to extract JSON from fenced code blocks. If no fenced JSON block is found, it falls back to a greedy \texttt{\{.*\}} regular expression, matching the extraction behavior used in WoW. Each predicted audit entry is then normalized to the expected schema, retaining the canonical fields \texttt{fieldname}, \texttt{newvalue}, \texttt{tablename}, and \texttt{oldvalue}. Finally, the normalized prediction is written to the task state as \texttt{predicted\_state\_json}.

\paragraph{Design motivation.}
This implementation intentionally exposes only a minimal query interface rather than a large collection of hand-engineered tools. The goal is to test whether an agent can recover transition logic from the same configuration artifacts that define the live enterprise system. By grounding its prediction in the current deployment, the discovery agent can adapt to tenant-specific rules and configuration changes without requiring retraining.

\paragraph{Pseudocode.}
The algorithm below consolidates the components described above into a single procedure: the outer autoregressive rollout, the inner ReAct loop with the \texttt{snow\_query} tool, and the deterministic post-processor.

\begin{taskbox}[Algorithm 1: Discovery Agent --- autoregressive state-prediction rollout]
{\small
\begin{alltt}
\textbf{Input:}  trajectory ((a_1, s_1), ..., (a_K, s_K)); compressed schemas
        Sigma, tool specs Tau; live instance configuration c.
\textbf{Output:} predicted per-step diffs hat_s[1:K], where each hat_s[t] is the
        set \{ (tablename, fieldname, oldvalue, newvalue) \}.

\textbf{1. Outer rollout (autoregressive over trajectory steps)}
hat_s := []
for t := 1, ..., K do
    state := InitState(a_t, hat_s, Sigma, Tau, step_index = t)
    msg   := DiscoveryAgent(state)
    hat_s.append( Postprocess(msg) )
return hat_s

\textbf{2. DiscoveryAgent: ReAct loop, single tool}
DiscoveryAgent(state):
    ctx := SystemPrompt + Render(state)
    while not FinalAnswer(msg) do
        msg := f_LLM(ctx)
        for (table, query, fields, limit, order_by) in msg.tool_calls do
            result := snow_query(table, query, fields, limit, order_by)
            ctx    := ctx + (msg, result)
    return msg

\textbf{3. Postprocess: extract \{audits: [...]\} from the final message}
Postprocess(msg):
    obj := ExtractFencedJSON(msg.content)            
    if obj is None then
        obj := GreedyJSONMatch(msg.content)          
    if obj is None or 'audits' not in obj then
        return \{ audits: [] \}
    return \{ audits: [
        \{tablename, fieldname, oldvalue, newvalue\} : a in obj.audits ] \}
\end{alltt}
}
\end{taskbox}

\section{CascadeBench: Failure Mode Analysis}
\label{app:cascade_analysis}
To characterize the gap between the oracle ceiling (prompted models 
with full business rules in context, 38--50\% IoU on CascadeBench) 
and perfect cascade prediction, we manually analyze two representative 
trajectories. Both are evaluated under the oracle condition: the 
model receives the schema, supporting data, and all relevant business 
rules in context, and is asked to predict the field-level cascade. 
We identify three recurring failure modes that account for the bulk 
of missed predictions even when context is complete.

\textbf{Failure patterns:}
\vspace{2pt}

\begin{patternbox}
\textbf{\textcolor{blue!70!black}{P1 \textbar\ Insert/creation blindness.}} 
When a business rule calls \texttt{gr.insert()}, the new record's fields 
produce 7--12 auditable changes. The model omits the insert entirely 
or predicts only 1--3 salient fields, missing the rest.
\end{patternbox}

\begin{patternbox}
\textbf{\textcolor{blue!70!black}{P2 \textbar\ Cascade fade-out.}} 
The model traces BR1--BR2 at 75--85\% recall but drops to 4--11\% 
for business rules at execution order $\geq 400$. Entire tables 
produced by deep-cascade rules are absent from predictions.
\end{patternbox}

\begin{patternbox}
\textbf{\textcolor{blue!70!black}{P3 \textbar\ Single-record assumption.}} 
When a business rule iterates a result set (\texttt{while (gr.next())}) 
or issues multiple inserts, the model predicts effects on exactly 
one record per table, missing parallel updates and sibling inserts.
\end{patternbox}

\vspace{4pt}
These are reasoning patterns, not retrieval patterns: the rules and 
supporting context are present in the prompt. They characterize a 
ceiling on what any context-fed approach (prompted oracle or 
perfect-retrieval discovery agent) can achieve without explicitly 
training the model to compose multi-step rule cascades.

\begin{taskbox}[Trajectory 1: Accounts Payable --- Invoice Match Status]
\textbf{Action.} \texttt{UPDATE} on \texttt{u\_ap\_vendor\_invoice} 
setting \texttt{u\_match\_status = 3} (Fully Matched).

\medskip
\textbf{Cascade structure.} A 6-rule cascade spans vendor invoices, 
purchase orders, line items, approval requests, and payment 
disbursements.

{\small
\begin{verbatim}
UPDATE vendor_invoice.match_status -> 3
|
+-- BR1 [order 100] vendor_invoice, purchase_order      [predicted well]
+-- BR2 [order 200] purchase_order, INSERT approval     [predicted well]
+-- BR3 [order 200] vendor, po_line_items(s)            [partial - P3]
+-- BR4 [order 400] po_line_items(s)                    [partial - P3]
+-- BR5 [order 500]
|   |-- invoice_line_item(s)                            [missed - P1+P2+P3]
|   |-- INSERT payment_disbursement                     [missed - P1]
|   |-- approval_request(s)                             [missed - P2+P3]
+-- BR6 [order 600]
    |-- invoice_line_item                               [missed - P2]
\end{verbatim}
}

\textbf{Result.} 22 predictions, 15 correct (precision 68\%, recall 
34\% against 44 ground-truth audits). The model traced BRs 1--2 
accurately but missed the \texttt{u\_ap\_invoice\_line\_item} table 
entirely (17 missed changes from BRs 5--6). Multi-pass overwrites 
on \texttt{u\_gl\_account\_code} were captured only for the first 
pass.
\end{taskbox}

\begin{taskbox}[Trajectory 2: Contract Obligation --- Overdue Escalation]
\textbf{Action.} \texttt{UPDATE} on \texttt{u\_cot\_obligation} 
setting \texttt{u\_status = 5} (Overdue) and \texttt{u\_priority = 4} 
(Critical).

\medskip
\textbf{Cascade structure.} A 5-rule cascade spans obligations, 
contracts, counterparties, escalations, audit logs, and reviews.

{\small
\begin{verbatim}
UPDATE obligation (status -> 5, priority -> 4)
|
+-- BR1 [order 100]                                     [predicted well]
|   INSERT escalation (7 fields)
|   INSERT audit_log #1 (11 fields)
|   UPDATE contract (3 fields)
+-- BR2 [order 200] contract overwrite                  [predicted well]
+-- BR3 [order 200]                                     [missed - P2]
|   counterparty fields, INSERT observer counterparty
+-- BR4 [order 400] observer defaults                   [missed - P2]
+-- BR5 [order 500]                                     [missed - P2+P3]
    UPDATE escalations (nested while)
    INSERT audit_log #2
    UPDATE reviews (while)
\end{verbatim}
}

\textbf{Result.} 26 predictions, 14 correct (precision 54\%, recall 
29\% against 48 ground-truth audits). The model traced BRs 1--2 
accurately (86--100\% of their outputs) but predicted 0\% of outputs 
from BRs 3--5. It produced one audit log record where two exist, 
and entirely missed \texttt{u\_cot\_obligation\_review} (9 changes) 
and \texttt{u\_cot\_counterparty} (2 changes).
\end{taskbox}

\subsection{Aggregate Failure Analysis}
\label{app:failure_analysis}

\begin{table}[h]
\centering
\small
\caption{Recall by cascade depth across both trajectories. Early 
business rules (depth 1--2) are well covered; deep rules (depth 
$\geq$ 3) are nearly entirely missed.}
\label{tab:depth_recall}
\begin{tabular}{lcc}
\toprule
\textbf{BR depth} & \textbf{AP Invoice} & \textbf{COT Obligation} \\
\midrule
Depth 1--2 (order 100--200) & 12/16 (75\%) & 17/20 (85\%) \\
Depth $\geq$ 3 (order $\geq$ 400) & 3/28 (11\%) & 1/28 (4\%) \\
\bottomrule
\end{tabular}
\end{table}

\begin{table}[h]
\centering
\small
\caption{False-negative attribution by failure pattern. Many false 
negatives carry multiple tags.}
\label{tab:fn_attribution}
\begin{tabular}{llcc}
\toprule
& \textbf{Description} & \textbf{AP} & \textbf{COT} \\
\midrule
P1 & Insert/creation-phase audits missed & 13 & 15 \\
P2 & Cascade fade-out (depth $>$ 2) & 19 & 20 \\
P3 & Single-record assumption & 5 & 8 \\
Other & Boolean coercion, ref-key, timestamp, value & 5 & 5 \\
\bottomrule
\end{tabular}
\end{table}

\paragraph{P1: Under-prediction of record creation.} A single missed 
insert produces 7--12 false negatives at once. Creation-phase audits 
(\texttt{old\_value = ""}) are recalled at 24--27\%, roughly half 
the rate of update audits (36--47\%), confirming the model is 
systematically weaker at predicting record creation than record 
modification.

\paragraph{P2: Cascade coverage drops after the first two rules.} 
Table~\ref{tab:depth_recall} shows the model simulates rule scripts 
accurately for the first 1--2 hops but does not sustain this across 
longer chains. Entire tables disappear from predictions, and 
multi-pass overwrites are half-predicted: when two rules write the 
same field sequentially, the model captures only the first write.

\paragraph{P3: Single-record assumption.} The model treats each table 
as having a single ``affected record.'' AP BRs 3--5 use 
\texttt{while (gr.next())} to iterate all PO line items, invoice 
line items, and approval requests; the model predicted changes to 
at most one record per table. COT BR3 updates the existing contract 
counterparty and inserts a new observer counterparty record; the 
model predicted only the update, not the sibling insert.

\paragraph{Implications.} These failure modes are reasoning 
bottlenecks of the underlying model, not retrieval failures: they 
persist when all rules are provided in the prompt. Closing the gap 
to perfect cascade prediction therefore requires training the model 
to compose rule executions, not just to retrieve them. This is the 
direction of the trained discovery agent proposed in 
\S\ref{sec:rung3}.
\section{Fine-tuning details}
\label{app:finetuning_details}
We fine-tune three recent open-weight language models with strong agentic benchmark performance: Qwen-3.5-27B~\citep{qwen3.5}, Qwen-3.6-27B~\citep{qwen3.6-27b}, and Gemma-4-31B-it~\citep{google_gemma_model_card}. We apply LoRA~\citep{hu2022lora} with rank 16 and $\alpha=32$ to the state-transition tuples from \S\ref{sec:gym}.

All models are trained for 2 epochs with a global batch size of 32 using AdamW ($\beta_1=0.9$, $\beta_2=0.95$, weight decay $0.01$) and a cosine learning rate schedule with 10\% linear warmup; learning rates are selected via a held-out validation set: $1\times10^{-4}$ for Qwen and $2\times10^{-4}$ for Gemma-4-31B-it. During fine-tuning, the maximum sequence length is set to 32k tokens for Qwen models and 12k tokens for Gemma-4-31B-it.
\section{Glossary}
\label{app:glossary}

The following terms describe enterprise platform concepts referenced throughout the paper.

\begin{description}[style=nextline, leftmargin=1.5em, labelindent=0em]

\item[Business Rule] (ServiceNow: \emph{Business Rules or BRs}) A server-side script that executes automatically when a record is created, updated, or deleted. Each rule has a trigger condition, an execution phase, and a numeric priority that determines firing order relative to other rules on the same table.

\item[Cascade] A chain reaction in which one business rule's output triggers another rule, which may trigger further rules. A single user action can produce cascades spanning multiple tables and dozens of intermediate rule firings.

\item[Configuration database] (ServiceNow: \emph{CMDB --- Configuration Management Database}) A structured repository of managed assets (servers, applications, network devices, software services) and their dependency relationships. Changes to one asset can propagate effects to all dependent assets.

\item[Execution phase] (ServiceNow: \emph{Business Rule ``When'' field}) The stage at which a business rule fires relative to the database operation: \emph{before} (can modify the record before it is written), \emph{after} (runs once the write is committed), \emph{async} (runs in a background thread), or \emph{display} (runs when the record is loaded for viewing).

\item[Instance] A single deployed installation of the enterprise platform, configured with its own business rules, access policies, and organizational structure. Different customers operate different instances with different configurations.

\item[Instance configuration ($c$)] The full collection of business rules, workflow definitions, approval policies, SLA definitions, and access control policies deployed on a particular instance. This is what makes the transition function instance-specific.

\item[Service-level agreement (SLA)] (ServiceNow: \emph{contract\_sla} and \emph{task\_sla}) A policy defining response and resolution time targets for tasks. When an SLA's start condition is met, the platform creates a timer record that tracks elapsed time and can schedule escalations.

\item[Access control policy] (ServiceNow: \emph{ACL --- Access Control List}) A rule governing which users or roles can read, write, or delete records on specific tables or fields. These can silently block or modify the effect of agent actions.

\item[Audit log] (ServiceNow: \emph{sys\_audit}) A platform-maintained record of every field-level change, including the old value, new value, timestamp, and the identity of the actor. This is the ground-truth source for capturing state transitions $(s_t, a_t, s_{t+1})$.

\end{description}



\end{document}